\documentclass{article}

\usepackage{arxiv}

\usepackage[utf8]{inputenc} % allow utf-8 input
\usepackage[T1]{fontenc}    % use 8-bit T1 fonts
\usepackage{url}            % simple URL typesetting
\usepackage{booktabs}       % professional-quality tables
\usepackage{amsfonts}       % blackboard math symbols
\usepackage{nicefrac}       % compact symbols for 1/2, etc.
\usepackage{microtype}      % microtypography
\usepackage{lipsum}		% Can be removed after putting your text content
\usepackage{graphicx}
\usepackage{fontawesome5}
\usepackage[numbers, sort, compress]{natbib}

\usepackage{doi}
\usepackage[normalem]{ulem}
\usepackage{hyperref}
\usepackage{xcolor}         % colors
\usepackage{amssymb}        % for symbols
\usepackage{multirow}
\usepackage{amsmath}

\usepackage{todonotes}
\usepackage{ulem}

\title{JetSeg: Efficient Real-Time Semantic Segmentation Model for Low-Power GPU-Embedded Systems}

\author{
  Miguel Lopez-Montiel\textsuperscript{1}\href{https://orcid.org/0000-0001-5367-9801}{\includegraphics[scale=0.10]{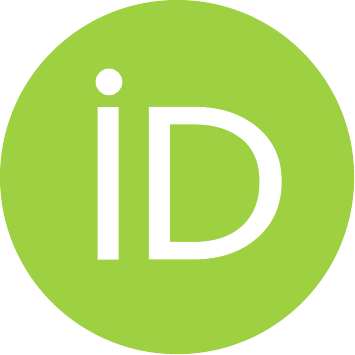}},
  Daniel Alejandro Lopez\textsuperscript{1}\href{https://orcid.org/0000-0002-3546-6762}{\includegraphics[scale=0.10]{orcid.pdf}},
  Oscal Montiel\textsuperscript{$*$, 1}\href{https://orcid.org/0000-0002-7060-9204}{\includegraphics[scale=0.10]{orcid.pdf}}\\
  \texttt{milopez@citedi.mx}, \texttt{dlopez@citedi.mx}, \\ \texttt{$^*$Corresponding author:~oross@ipn.mx}\\
  Department of Intelligent Systems\\
  Instituto Politécnico Nacional, CITEDI-IPN\textsuperscript{1}\\
  Av. Instituto Polit\'ecnico Nacional No. 1310, Nueva Tijuana C.P. 22435\\
}

\renewcommand{\shorttitle}{Real-time Semantic Segmentation Model for Embedded Systems}

\hypersetup{
pdftitle={A template for the arxiv style},
pdfsubject={q-bio.NC, q-bio.QM},
pdfauthor={David S.~Hippocampus, Elias D.~Striatum},
pdfkeywords={First keyword, Second keyword, More},
}

\begin{document}
\maketitle

\begin{abstract}
Real-time semantic segmentation is a challenging task that requires high-accuracy models with low-inference times. Implementing these models on embedded systems is limited by hardware capability and memory usage, which produces bottlenecks. We propose an efficient model for real-time semantic segmentation called JetSeg, consisting of an encoder called JetNet, and an improved RegSeg decoder. The JetNet is designed for GPU-Embedded Systems and includes two main components: a new light-weight efficient block called JetBlock, that reduces the number of parameters minimizing memory usage and inference time without sacrificing accuracy; a new strategy that involves the combination of asymmetric and non-asymmetric convolutions with depthwise-dilated convolutions called JetConv, a channel shuffle operation, light-weight activation functions, and a convenient number of group convolutions for embedded systems, and an innovative loss function named JetLoss, which integrates the Precision, Recall, and IoUB losses to improve semantic segmentation and reduce computational complexity. Experiments demonstrate that JetSeg is much faster on workstation devices and more suitable for Low-Power GPU-Embedded Systems than existing state-of-the-art models for real-time semantic segmentation. Our approach outperforms state-of-the-art real-time encoder-decoder models by reducing 46.70M parameters and 5.14\% GFLOPs, which makes JetSeg up to 2x faster on the NVIDIA Titan RTX GPU and the Jetson Xavier than other models. The JetSeg code is available at \href{https://github.com/mmontielpz/jetseg}{https://github.com/mmontielpz/jetseg}.
% {\color{blue}Metrics indicate that JetSeg is 1.15\% and 1.05\% better for the Camvid dataset. In addition, it is one of the first models capable of running in real-time on a Jetson Nano 4GB device while maintaining 70\% mIoU without model compression.} 

\end{abstract}

% keywords can be removed

\section{Introduction}
Semantic segmentation is one of the critical computer vision tasks for autonomous systems. It requires a complete understanding of the analyzed scene to recognize and divide each object comprising the image, video, or 3D data. Semantic segmentation aims to classify each pixel into its corresponding class label to understand the image's objects. The significance of a scene depends heavily on the destined application that may range from topics such as human-machine interaction \cite{Lepetit}, medical imaging \cite{UNet}, and autonomous driving methods \cite{Geiger2012}. For instance, systems directed toward autonomous vehicles must be able to segment roads and detect objects and traffic signs, which must operate under real-time conditions and, often, limited computing capabilities.

Since the mid-2010s, most semantic segmentation methods have used some variation of deep convolutional neural networks. Early networks such as fully convolutional networks~\cite{FCN}, SegNet~\cite{SegNet}, and U-Net \cite{UNet} revolutionized the capabilities of segmentation research. Nonetheless, these methods require high computational resources, which is unsuitable for real-time embedded system applications. 

In robotics, real-time semantic segmentation must reach breakneck running speeds for quick interaction and responses to aid decision-making. One of the earliest networks for real-time semantic segmentation was ENet~\cite{ENet}, which implemented a bottleneck module in a residual layer that reduced it to up to 1\% of SegNet~\cite{SegNet} parameters. %ICNet and ContextNet incorporated high-level label guidance through a cascaded network that uses an image pyramid for the inputs~\cite{ICNet, ContextNet},

Other proposals rely on pyramid feature extraction for real-time segmentation, such as FPENet \cite{FPENet} that produces an efficient feature pyramid for multi-scale context encoding, ESPNet that adopts efficient spatial pyramid dilated convolutions \cite{ESPNet}, and DFANet \cite{DFANet} that mixes multibranch and spatial pyramid pooling, and reduces complexity by reusing features enhancing feature representation. 

Although these methods can achieve real-time inference speed, the significant loss of features hinders their performance considerably, sacrificing accuracy for efficiency. Due to this, few works have attempted to obtain a favorable trade-off between high-quality results and low-inference times for embedded system applications. One of these models is ADSCNet \cite{ADSCNet}, a lightweight Asymmetric depthwise separable convolution network that connects sets of dilated convolutional layers using dense dilated convolution connections; this model achieves an mIoU of 67.5\% on the Cityscapes dataset at 76.9 FPS while performing 21.1 MFLOPS and having around 20K parameters while running on a GTX 1080 Ti GPU. Similarly, ESPNet V2 \cite{ESPNetV2} implements group pointwise and depth-wise dilated separable convolutions for representation learning. Experimentation was carried out on a GTX 1080 Ti GPU and an NVIDIA Jetson TX2, reaching an mIoU of 66.2-66.4\% by performing 2.7 B FLOPs for the Cityscapes dataset and 67-68\% mIoU with 0.76 B FLOPs respectively.
In recent work, MFENet~\cite{MFENet} uses modules for Spatial and Edge Extraction with a Laplace Operator to improve low-level feature extraction, a Context Boost Module for high-level features, and a Selective Refinement Module that combines their information. This network obtains an mIoU of 76.7\% on the Cityscape dataset performing 12.5 GLOPs at a speed of 47 FPS, and 73.5\% on the CamVid dataset when executed on an NVIDIA Titan Xp card. 
DSANet~\cite{DSANet}, a two-branch model that employs channel split and shuffle to reduce computation and attain high accuracy. Its dual attention and channel attention modules help with pixel-wise label prediction using multi-level feature maps simultaneously. The experiments on a GTX 1080 GPU for the CamVid and Cityscapes datasets achieved mIoU of 69.9\% and 71\% at speeds of 75.4 FPS and 34.04 FPS, respectively.
 MiniNetv2 \cite{MiniNet} was developed for low-power systems, improving the MiniNet efficient real-time semantic segmentation. It attains an mIoU of 70.4\% performing 12.89 GFLOPs and 6.45 GMACs at 50 FPS, having 0.52 M parameters and a memory footprint of 2.02 MB for a resolution of 1024 x 512 of the CityScapes benchmark while running on an NVIDIA Titan Xp card. Furthermore, several works regarding embedded system implementation show promising results when compared to workstation devices, such as ThunderNet~\cite{ThunderNet} is an efficient and lightweight network with a minimal ResNet18 backbone that unifies PSPNet pyramid pooling and a decoding phase that achieves an mIoU of 64\% at 96.2 FPS on a Titan Xp GPU and 20.9 FPS on a Jetson TX2. Another instance is WFDCNet, a network of full-dimensional continuous separation convolution modules (FCS) and lateral asymmetric pyramid fusion modules (LAPF) that enable it to attain high accuracy without hindering inference speed, reaching 73.7\% of mIoU having 5.7 GFLOPs at 102.6 FPS and 0.5 M parameters on an RTX 2080 Ti GPU for the Cityscapes dataset while achieving a speed of 17.2 FPS on a Jetson TX2. EFNet \cite{EFNet}, a modified version of ShuffleNet V2 , aims to further improve accuracy-speed trade-off by reducing its downsampling operations, stride, and deconvolutions. Two thinner convolutions replace its large convolution channel, and the final downsampling and upsampling channels are substituted by more elaborate modules that boost efficiency. As a result, it reaches an mIoU of 68\% at 99 FPS and 0.18 M parameters on an RTX 3090 GPU for the Cityscapes dataset. However, when run on a Jetson Nano, the model achieves similar accuracy results but significantly lower inference speeds of 0.9 FPS. Nevertheless, it was tested on a novel memristor-based computing-in-memory accelerator to demonstrate its efficiency on embedded systems further, obtaining 40.8 FPS and comparable accuracy results. 

%One of the constant challenges in semantic segmentation is integrating global context information with fine spatial detail in search of attaining real-time models \cite{Garcia-garcia2018}.

Additionally to the trade-off between speed and accuracy of current real-time semantic-segmentation models, there is limited research on execution time and memory footprint for real-time models. Therefore, due to the necessities mentioned above, our work focused on developing an efficient model capable of achieving real-time semantic segmentation in low-power embedded devices; specifically, we contribute to state-of-the-art with the following:

\begin{itemize}

    \item We introduce a novel, light, and efficient architecture called JetSeg that can compete with state-of-the-art faster and bigger models, outperforming them in most cases. JetSeg contains the novel encoder JetNet, the novel block JetBlock, the novel operation JetConv, and features an original loss function named JetLoss, which contributions will be explained.
    
   % \item We introduce a novel, light, and efficient architecture called JetSeg that can compete with state-of-the-art faster and bigger models, outperforming them in most cases. JetSeg contains the novel blocks JetConv, JetNet, JetBlock, and JetLoss, which contributions will be explained next. 
    
    \item JetConv is a novel convolutional operator designed for lightweight and efficient feature extraction in real-time, low-power embedded systems. It enhances spatial information representation, increases the receptive field, and captures long-range dependencies without adding parameters. In addition, by combining dilated asymmetric and non-asymmetric convolutions, JetConv maintains spatial symmetry and enables balanced feature extraction in all directions. 

    \item JetNet is an encoder that can extract information without compromising speed. It consists of convolutional blocks comprised of JetConvs. By combining dilated asymmetric depthwise convolutions and non-asymmetric traditional convolutions, the encoder can acquire large amounts of context information on fine and large dependencies without drastically increasing the number of parameters.

    \item JetBlock is an efficient unit for feature map extraction, achieving a balance between inference time, memory usage, and model abstraction. It consists of multiple stages for optimal performance, estimates the number of group convolutions to prevent overfitting, and improves generalization. In addition, batch normalization and lightweight activation functions like TanhExp maintain representation while minimizing inference time. 

    \item JetLoss is a novel loss function for improved semantic segmentation. It combines Precision, Recall, and IoUB losses, enhancing overall performance. Adaptive weightings based on pixel counts per class are included to further enhance effectiveness by focusing on challenging classes. As a result, JetLoss enables faster convergence, efficient training, and reduced computational requirements. Experimental results on the CamVid \cite{camvid} benchmark dataset show its edge over traditional loss functions, positioning JetLoss as a strong competitor in semantic segmentation.  
\end{itemize}
\section{JetSeg}
According to the revised related works, most real-time semantic segmentation models are unsuitable for low-power embedded devices. Therefore, our work was to design an efficient and lightweight model capable of running on resource-constrained devices, such as the NVIDIA Jetson Nano and NVIDIA Jetson AGX, due to most embedded devices possessing GPU as hardware accelerators \cite{Shuvo2023}.

\begin{figure*}[ht!]
    \centering
    \includegraphics[width=\textwidth]{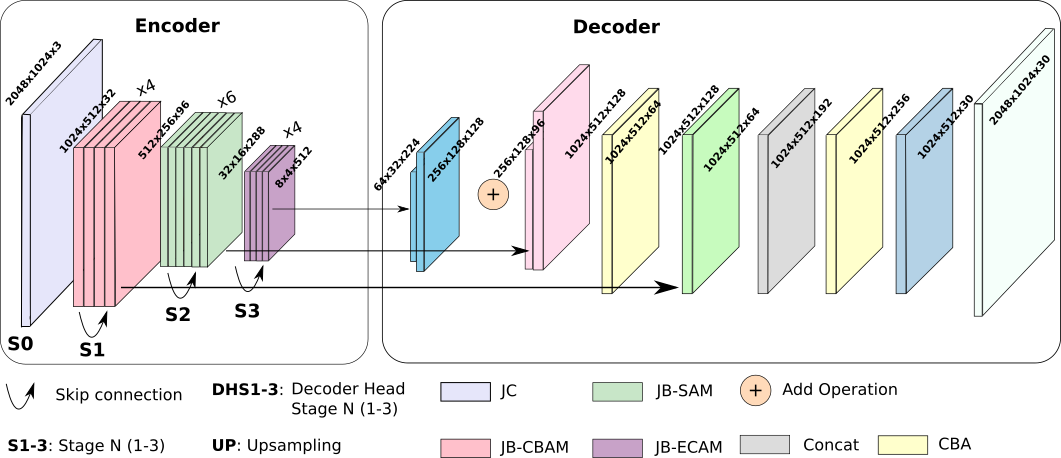}
    \caption{JetSeg framework: Our proposed JetSeg follows the encoder-decoder architecture. The encoder component is JetNet, while the decoder is based on RegSeg.}
    \label{fig:Architecture}
\end{figure*}

Our proposed model JetSeg, see Figure \ref{fig:Architecture}, is an efficient real-time semantic segmentation model for low-power GPU embedded systems, specifically, NVIDIA Jetson devices. This model comprises two architectures: JetNet, our proposed encoder, and an enhanced version of the RegSeg~\cite{gao2021} decoder; hence the name JetSeg for our model. JetSeg was designed for memory and operation efficiency while preserving a competitive mIoU and FPS to the state-of-the-art (SOTA) models for low-power real-time semantic segmentation without model optimization or compression. 

%%% Figure 1

The JetSeg model has three configurations, one for the workstation, one for the NVIDIA Jetson AGX, and one for the NVIDIA Jetson Nano. Input features are first expanded and later reduced when nearing the output layers, which are processed in different phases depending on the stage they are in. Four cases dictate the stage and operations applied to the input features. S0 and S1 focus on high-level feature extraction, using three main operations: channel shuffle to reduce the number of parameters; JetConv, which captures low and high-level features without increasing model complexity; and a convolutional block (CBAM) \cite{woo2018cbam} that enables the model to learn more discriminative and robust features. S0 also includes residual layers just as S2; however, S2 replaces the CBAM with the SAM \cite{woo2018cbam}, which allows the model to capture long-range dependencies and enhances spatial awareness. S3 presents half the residual layers compared to S2 and also includes an Efficient Channel Attention Module \cite{wang2020ecanet} that highlights relevant channel-specific information.
\subsection{JetConv}
\begin{figure}[ht!]
  \centering
  \includegraphics[scale=0.55]{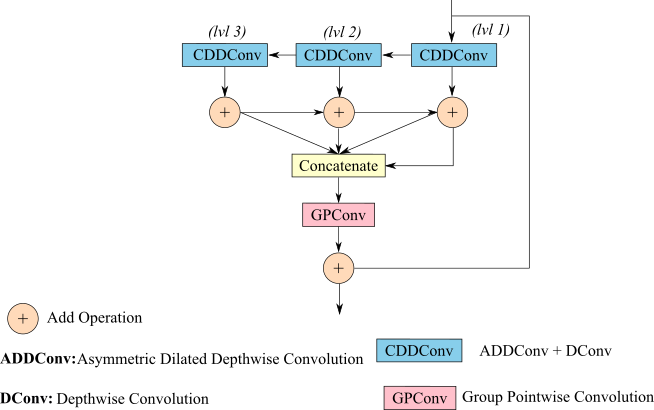}
  \caption{JetConv is based on Standard/non-Asymmetric Dilated Convolution by the three levels that comprise JetBlock.}
  \label{fig:JetConvs}
\end{figure}

The JetConv novel operator is for lightweight and efficient feature extraction in real-time low-power embedded systems. It was inspired by the EESP (Extremely Efficient Spatial Pyramid) module \cite{ESPNetV2}. The dilated Depthwise Convolutions have been improved by introducing Combinated Dilated Depthwise Convolution (CDDC), which combines dilated asymmetric convolution followed by a non-asymmetric and non-dilated convolution. This method enables a more effective and balanced spatial information representation. The receptive field is also significantly increased, which allows capturing long-range dependencies without increasing the total number of parameters and balancing between power and computational efficiency. Asymmetric convolution limitations are compensated through non-asymmetric convolutions, which preserve spatial symmetry and allow equal feature extraction in all directions. This symmetry is beneficial when dealing with objects lacking directional properties and ensures balanced information among spatial dimensions. JetConv improves contextual understanding without sacrificing spatial symmetry or robustness by integrating the two types of convolutions, allowing it to capture long-range dependencies and global spatial patterns critical for semantic segmentation. Figure \ref{fig:JetConvs} shows the structure of the JetConv.
This balanced approach benefits objects without directional properties, ensuring comprehensive spatial information. JetConv achieves improved contextual understanding, capturing critical long-range dependencies and global spatial patterns for semantics while preserving spatial symmetry and robustness. Figure \ref{fig:JetConvs} is a residual block composed by five layers. In the first layer, there are three CDDConv blocks; each CCDConv block is composed of two different convolution operations. The first JetConv configuration consists of a depthwise convolution (DConv) \cite{depthwiseConv}, ADDConv (Asymmetric Dilated Depthwise Convolution) \cite{DABNet}, and the novel convolution operation name ADDConv; they can configure different size kernels with and without dilatation rate, they are: $3 \times 3$, $5 \times 5$ and $7 \times 7$ for non-asymmetric, and for dilated asymmetric ($3 \times 1$ + $1 \times 3$), ($5 \times 1$ + $1 \times 5$) and ($7 \times 1$ + $1 \times 7$). The second is an additive layer that adds the features obtain in the first layer, depending on the number of CDDConv blocks we have. The objective of the third layer is to concatenate the result of the second layer, i.e., we stack the obtained results of each CCDConv block to combine the feature extraction of each configuration of CCDConv. The GPConv in the fourth layer red reduces the overfitting, memory usage, and inference time. The objective of the last layer is to perform feedback of the GPConv result to the first layer to complete the residual calculus. The JetConv block allows three basic modifiable configurations: a) the first one contains one CDDConv block in the first, layer followed by the four layers shown in Figure \ref{fig:JetConvs}, b) the second configuration adds a CCDConv block to the first configuration, c) the third one adds a third CCDConv block the second configuration.

\subsection{JetBlock}
\begin{figure}[ht!]
  \centering
  \includegraphics[scale=0.55]{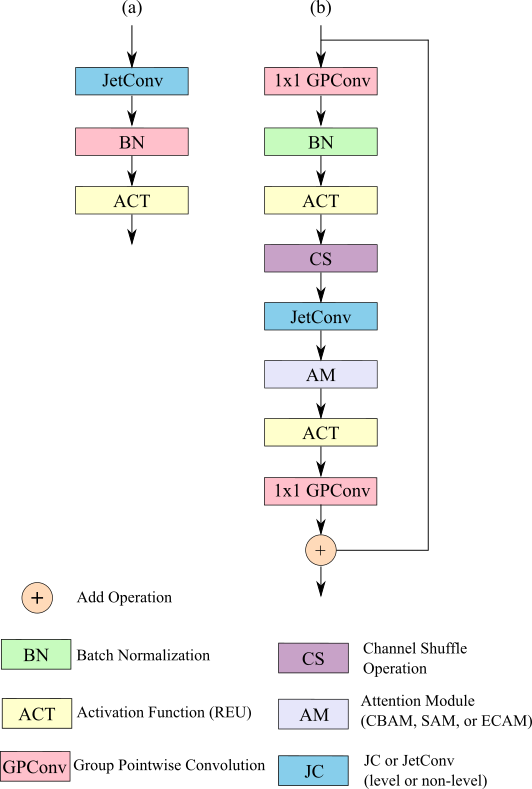}
  \caption{Overview of JetNet Blocks: (a) Input block, and (b) Subsequent blocks that comprise JetNet.}
  \label{fig:JetBlocks}
\end{figure}

JetBlock is a new efficient unit for feature map extraction comprised of several stages that allow it to achieve an equilibrium between inference time, memory usage, and model abstraction. First, the convenient number of group convolutions is estimated, preventing overfitting and improving model generalization. Group convolutions are followed by batch normalization and activation functions suited for lightweight architectures, being TanhExp used to reduce inference time while maintaining capture and representation. The second stage focuses on computational efficiency through the channel shuffle operator, which optimizes memory access patterns, enhances hardware resource management, promotes feature diversity through layer interconnections, and speeds up computation. The third stage extracts high-level features using the JetConv layer and distinguishes between the important elements with the attention module layer. Afterward, the block finishes with an activation function and a grouped pointwise operator, reducing the computational cost and memory requirements. The composition of the JetBlocks is defined in Figure \ref{fig:JetBlocks}.
%%% Figure 3

\subsection{JetNet}
JetNet is an encoder architecture that heavily relies on a set of efficient convolutional blocks called JetConvs. It has four stages as S0 to S3, as shown in Figure \ref{fig:Architecture}. The Stage S0 is composed of one configurable JetBlock followed by a Batch Normalization (BN) and an Activation Function REU (Rectified Exponential Units) \cite{reu} as shown in \ref{fig:JetBlocks}(a). The Stage S1 is a residual block composed of four configurable JetBlocks as shown in \ref{fig:JetConvs}(b); in this stage, the four blocks are similar in structure. Its first layer is composed of a GPConv of $1 \times 1$ that expands the number of input channels in order to extract more features, followed by BN,  REU, Channel Shuffle (CS), JetConv, Attention Module (AM), REU and $1 \times 1$ GPConv in order to reduce the number of features. Note that the AM layer is a CBAM module. Stage S2 is similar to stage S1; the difference is the configuration, particularly in the AM module, and the number of features used.
We used only the spatial attention module in the AM module because the image size was reduced. Stage S3 is similar to stage S2; the block of this stage where reduced to improve the processing speed of JetNet. In this stage, the attention module was also changed; it used ECAM instead of SAM.

The main goal of the JetConv layer is to extract feature maps with as few operations as possible. The JetConvs are the main blocks for the two JetNet block variations; see Figure~\ref{fig:JetBlocks}. The input block consists of a JetConv, a batch normalization layer, and an activation function. The subsequent stages of JetSeg are more intricate. The beginning and end of these blocks are similar to the input block; there, we replace the JetConv with a $1 \times 1$ convolution to reduce computational complexity. After that, we continue with channel shuffle as proposed in \cite{zhang2017shufflenet} followed by another JetConv, then an attention module, an activation function, and we finish by a $1 \times 1$ convolution. The attention module can be any of the following: a spatial attention module (SAM)~\cite{woo2018cbam}, a CBAM, or an efficient channel attention module (ECAM)~\cite{gao2021}. Finally, we use a skip connection to concatenate the residual for each block. 

\subsection{JetLoss}

The JetLoss function is a novel loss function inspired by the work of Tian \cite{Tian2021}, designed to enhance model performance in semantic segmentation. JetLoss integrates three loss functions into one, combining elements from Precision \cite{Flach2015}, Recall \cite{Tian2021}, and IoUB \cite{IoUB} losses to improve the overall performance, particularly in terms of recall.

To further improve the effectiveness of JetLoss, adaptive weightings based on the number of pixels per class, following Cui's approach \cite{Cui2019}, are incorporated. These adaptive weightings assign importance to each class during training, enabling the model to focus on challenging classes and achieve better overall performance. As a result, using JetLoss promotes faster model convergence, efficient training, and reduced computational resource requirements. Moreover, it ensures a fair evaluation of the semantic segmentation model's performance.

The experimental results on the CamVid \cite{camvid} benchmark dataset demonstrate the advantages of using JetLoss compared to conventional loss functions, highlighting its potential as a robust contender in semantic segmentation loss function proposals.

The components of the JetLoss function are described by the following set of equations \ref{eq:jetloss}:

\begin{align}
    R_{loss} &= \sum_{c=1}^{c}\frac{TP_{c}+\epsilon}{TP_{c}+FN_{c}+\epsilon} \times W_c\nonumber \\
    P_{loss} &= \sum_{c=1}^{c}\frac{TP_{c}+\epsilon}{TP_{c}+FP_{c}+\epsilon} \times W_c \nonumber \\
    B_{loss} &= \frac{{\sum_{i=1}^{N} TP_{b_i}}}{{\sum_{i=1}^{N} (TP_{b_i} + FP_{b_i} + FN_{b_i})}} \times W_c \nonumber\\
    JetLoss &= R_{loss} + P_{loss} + B_{loss} \nonumber\\
    \label{eq:jetloss}
\end{align}
In the equations above, $R_{loss}$ represents the recall loss, $P_{loss}$ represents the precision loss, and $B_{loss}$ represents the IoUB loss. $TP_{c}$, $FN_{c}$, $FP_{c}$, and $W_c$ denote true positives, false negatives, false positives, and class-specific weights, respectively. The overall JetLoss is computed as the sum of the three component losses.

\section{Experiments and results}
The following metrics are used to evaluate JetSeg and are generally employed to gauge the performance of real-time semantic segmentation models.
The computational complexity is essential when working with limited hardware resources; hence the FLOPs metric is fundamental for evaluating a CNN model. It measures the total number of arithmetic computations performed in a process, emphasizing additions and multiplications. According to authors such as \cite{Wang2021}, several factors play a part in FLOPs, for instance, the expense in the computation of convolutions ($F_{conv}$), fully connected layers ($F_{fc}$), activation functions ($F_{act}$), as well as batch normalizations ($F_{bn}$). Equation \ref{eq:flops} shows the total FLOPs formula broken down,
%% Equation 1
\begin{align}
F_{conv} &= \frac{K_w \times K_h \times H_{in} \times W_{in} \times C_{in} \times C_{out}}{s^2 \times g}\nonumber \\
F_{pool} &= C_{in} \times H_{out} \times W_{out} \times K_w \times K_h \nonumber \\
F_{fc} &= C_{in} \times H_{in} \times W_{in} \times H_{out} \times W_{out} \nonumber \\
F_{bn} &= 4 \times C_{in} \times H_{out} \times W_{out} \nonumber \\ 
F_{act} &= C_{in} \times H_{in} \times W_{in} \nonumber \\
FLOPs &= F_{conv} + F_{pool} + F_{fc} + F_{act} + F_{bn}
\label{eq:flops}
\end{align}
where $K_w$, $K_h$, $H_{in}$, and $W_{in}$ represent the kernel size, input height, and input width of the convolution layer, respectively; input and output channels are represented by $C_{in}$ and $C_{out}$; $H_{in}$, $W_{in}$, $H_{out}$ and $W_{out}$ represent the input height, input width, output height and output width of the fully connected layer; activation function cost is represented by $F_{act}$ and computed with the number of activation functions in the input tensor given by $C_{in} \times H_{in} \times W_{in}$, as well as the number of operations needed per activation; and $F_{bn}$ represents batch normalization and is calculated based on the number of activations in the input tensor, which is given by $C_{in} \times H_{in} \times W_{in}$. 
The fundamental metric to evaluate semantic segmentation performance is the mIoU \cite{IoUB}, which refers to the total percentage of pixels belonging to the ground truth class that the model was able to classify, meaning the overlapping degree between the ground truth and model prediction. Equation \ref{eq:mIoU} describes the computation.
\begin{align}
    mIoU &= \frac{1}{|C|}\sum_{c} \frac{\sum_{i \in P}TP_{ic}}{\sum_{i \in P}TP_{ic} + FP_{ic} + FN_{ic}} \label{eq:mIoU}
\end{align}
The variables $TP_{c}$, $FP_{c}$, and $FN_{c}$ represent the counts of True Positive, False Positive, and False Negative pixels, respectively, for class $c$. The set $D = {1, 2, ..., C}$ refers to all the categories present in the dataset, while $P$ represents the set of all pixels within the dataset.

This section describes of the experimental setup and results analysis used to evaluate our proposed JetSeg model. We conducted our experiments using the widely recognized dataset CamVid, commonly used as a benchmark dataset for Autonomous Driving Semantic Segmentation. CamVid \cite{camvid} is a database for understanding road and driving scenes initially from five videos, which add to a total of 701 frames divided into 32 classes.

We conducted two phases of experiments: a classification evaluation and a real-time embedded system implementation. We utilized the CamVid dataset for both. A split of 67\% of the data for training (468 samples) and 33\% for testing (233 samples). Additionally, we divided the training set into 78\% (367 samples) for training and 22\% (101 samples) for validation.

The hardware used for training and evaluation consisted of an NVIDIA TITAN RTX GPU. We employed GPU-equipped embedded system boards to evaluate the real-time embedded implementation: the NVIDIA Jetson Nano and the NVIDIA Jetson AGX. To develop this work, we utilized a variety of software tools. Firstly, we employed PyTorch \cite{Paszke2019}, the currently most widely used framework for deep learning. To ensure the reproducibility of our experiments, we utilized Docker containers \cite{Merkel2014} for all our hardware, including both our workstation and NVIDIA Jetsons.\footnote{\href{https://catalog.ngc.nvidia.com/orgs/nvidia/containers/l4t-ml}{https://catalog.ngc.nvidia.com/orgs/nvidia/containers/l4t-ml}}

\begin{table}[!ht]
    \caption{SoTA Results for Real-time Semantic Segmentation on the CamVid dataset (non-Jetson Devices). $\ast$ represents unclear information in some columns. G refers to Giga.}
    \fontsize{9}{9}\selectfont
    \label{tab:sota-camvid-njetsons}%
    \centering
    \begin{tabular}{ccccccccc}
    \toprule
    \multirow{2}{*}{Model}      & Input                            & \multirow{2}{*}{Pretrained}    & Data                      & Model                 & Params               & FLOPs                & \multirow{2}{*}{FPS}\\
                                & Size                             &                                &  Aug                      & Opt                   & (G)                  &  (G)                &\\
    \midrule
    $\ast$LinkNet \cite{LinkNet}& $640 \times 360$                 & No                        &     No                     &   No                   & 11.5                 & 21.2                     & -\\
    $\ast$CGNet  \cite{CGNet}   & $480 \times 360$                 & No                       &     Yes                    &   No                  & 0.5                    & 6                          & -\\
    DSANet \cite{DSANet}        & $480 \times 360$                 & No                       &     Yes                    &   No                  & 3.47                  & 37.4                        & 75.3\\
    MFENet \cite{MFENet}        & Original                         & Yes                      &     Yes                    &   No                  & 3.91                  &   12.5                       & -\\

    JetSeg                & \multirow{2}{*}{$512 \times 512$} & \multirow{2}{*}{No}      & \multirow{2}{*}{No}       & \multirow{2}{*}{No}   & \multirow{2}{*}{0.00003}  & \multirow{2}{*}{1.125}   &\multirow{2}{*}{158}\\
    (Ours)  &                                  &                          &                           &                       &                       &                       &                     \\
    \bottomrule
    \end{tabular}
\end{table}%

The following section presents our evaluation of the JetSeg model in a real-time setting, showcasing the model's efficiency on embedded devices with limited resources. The evaluation was conducted on NVIDIA Jetson GPU-Embedded Devices using NGC Docker containers to ensure the reproducibility and generalizability of the experiments. For Jetson AGX Xavier, we utilized JetPack version $5.0.2$ and the NGC Docker container \texttt{nvcr.io/nvidia/l4t-ml:r35.1.0-py3}. For Jetson Nano 4GB, we used JetPack $4.6.1$ and the NGC Docker container \texttt{nvcr.io/nvidia/l4t-ml:r32.7.1-py3}. These configurations allowed us to effectively demonstrate the performance of JetSeg in real-time scenarios with constrained resources.

The experiments are divided into two categories: workstation and embedded devices. During the training phase, the model is trained from scratch for 15 epochs, without any pre-training, model compression, or data augmentation techniques. Table \ref{tab:sota-camvid-njetsons} presents the performance metrics of the model on the CamVid dataset with an input size of $512 \times 512$. The model achieves a considerably fast speed of 158 FPS, performing 1.125 G FLOPs with only 0.00003 G parameters. These results highlight that JetSeg is a compact and high-speed model for semantic segmentation.

When evaluated on embedded devices, the NVIDIA Jetson AGX demonstrates real-time performance, achieving a speed of 39.9 FPS. Despite its low hardware requirements, the Jetson AGX performs 1.125 G FLOPs with 0.00003 G parameters, making it suitable for real-time semantic segmentation in embedded system applications. Moreover, the model also runs successfully on the NVIDIA Jetson Nano, showcasing its lightweight nature and capability to operate on low-memory devices.

\section{Discussion}
This work proposes a novel model called JetSeg for the task of semantic segmentation for embedded devices in real-time, presenting an innovative, efficient, and lightweight architecture comprised of an encoder called JetNet, new model blocks named JetBlocks, original convolution operations called JetConv, and a new loss function called JetLoss, which allows it to run on embedded devices such as the NVIDIA Jetson AGX in real-time. This architecture lets the model extract low and high-level features while reducing computational complexity, maintaining spatial symmetry and balanced feature extraction thanks to the JetConv convolutional operator. JetNet combines dilated asymmetric depthwise convolutions and non-asymmetric traditional convolutions to extract context information from fine and large dependencies. JetBlock comprises JetNet and is an efficient unit for feature extraction, preventing overfitting while minimizing inference time. Lastly, JetLoss improves semantic segmentation performance by integrating three losses: Precision, Recall, and IoUB, which enhances overall performance through adaptive weighting based on pixel counts per class. This enables faster convergence and efficient training with reduced computational requirements. JetSeg can be executed in three configurations suited for workstations, such as the NVIDIA Titan RTX GPU, the NVIDIA Jetson AGX, and NVIDIA Jetson Nano embedded devices achieving high FPS, such as 148 FPS on the Titan RTX GPU and 39.1 FPS on the Jetson AGX, inference speeds suited for real-time implementation. The comparative experiment results showed that JetSeg outperformed the tested state-of-the-art models in inference speed up to 2$\times$ while executing less than 5$\times$ GFLOPs, as well as preserving a small number of parameters without model optimization, compression techniques, pretraining, or data augmentation.

%{\color{blue}
\section*{Authorship Note}
%The authors, \href{https://orcid.org/0000-0002-5718-9700}{\includegraphics[scale=0.06]{orcid.pdf} Yoshio Rubio} and \href{https://orcid.org/0000-0001-7129-1659}{\includegraphics[scale=0.06]{orcid.pdf} Cynthia Olvera}, are not included in this paper because the idea and project development are solely attributed to \href{https://orcid.org/0000-0001-5367-9801}{\includegraphics[scale=0.06]{orcid.pdf}\hspace{1mm}Miguel Lopez-Montiel} and contributors \href{https://orcid.org/0000-0002-3546-6762}{\includegraphics[scale=0.06]{orcid.pdf}\hspace{1mm}Daniel Alejandro Lopez} and \href{https://orcid.org/0000-0002-7060-9204}{\includegraphics[scale=0.06]{orcid.pdf}\hspace{1mm}Oscar Montiel}. While Yoshio Rubio and Cynthia Olvera were intended to contribute by collaborating on the writing and document review, time constraints prevented them from fulfilling these roles.}

The contribution for the realization of this paper is as follow: 
The idea and project development are attributed to Miguel Angel Lopez Montiel and collaborators Daniel Lopez Montiel and Oscar Montiel Ross. In the submitted paper to NeurIPS2023, the authors Yoshio Rubio and Cynthia Olvera were included to participate in the writing and document review; however, they could not participate due to time constraints preventing them from fulfilling these roles; hence, they do not appear in this paper.

\bibliographystyle{unsrtnat}
\bibliography{references}  %%% Uncomment this line and comment out the ``thebibliography'' section below to use the external .bib file (using bibtex) .





% \end{thebibliography}

\end{document}